\documentclass{article}

\usepackage{PRIMEarxiv}

\usepackage[utf8]{inputenc} 
\usepackage[T1]{fontenc}    
\usepackage{hyperref}       
\usepackage{url}            
\usepackage{booktabs}       
\usepackage{amsfonts}       
\usepackage{nicefrac}       
\usepackage{microtype}      
\usepackage{lipsum}
\usepackage{fancyhdr}       
\usepackage{graphicx}       
\graphicspath{{media/}}     

\usepackage{bm}
\usepackage{natbib}
\usepackage{adjustbox}
\usepackage{bbm}
\usepackage{amsmath}
\usepackage{makecell}
\usepackage{caption}

\title{KGBoost: A Classification-Based Knowledge Base Completion Method with Negative Sampling}

\author{
  Yun-Cheng Wang \\
  University of Southern California \\
  Los Angeles, USA\\
  \texttt{yunchenw@usc.edu} \\
   \And
  Xiou Ge \\
  University of Southern California \\
  Los Angeles, USA\\
  \texttt{xiouge@usc.edu} \\
   \And
  Bin Wang \\
  National University of Singapore \\
  Singapore\\
  \texttt{bwang28c@gmail.com} \\
   \And
  C.-C. Jay Kuo \\
  University of Southern California \\
  Los Angeles, USA\\
  \texttt{cckuo@sipi.usc.edu} \\
}

\begin{document}
\maketitle

\begin{abstract}
Knowledge base completion is formulated as a binary
classification problem in this work, where an XGBoost binary classifier
is trained for each relation using relevant links in knowledge graphs
(KGs). The new method, named KGBoost, adopts a modularized design and 
attempts to find hard negative
samples so as to train a powerful classifier for missing link
prediction.  We conduct experiments on multiple benchmark
datasets, and demonstrate that KGBoost outperforms state-of-the-art
methods across most datasets. Furthermore, as compared with
models trained by end-to-end optimization, KGBoost works well under the low-dimensional setting so as to allow a smaller model size.
\end{abstract}


\section{Introduction}\label{sec:introduction}

Knowledge graphs (KGs) are structured representations of factual
triples. A triple, denoted by $(h, r, t)$, describes the relationship,
$r$, between the head entity, $h$, and the tail entity, $t$. Real-world
KGs, such as Freebase \citep{bollacker2008freebase}, WordNet
\citep{miller1995wordnet} and NELL \citep{carlson2010toward}, contain
millions of triples. Yet, most KGs still suffer from the problem of
incompleteness, i.e., missing relation links between entities. For
example, 71\% of people in Freebase do not have the place of birth
information \citep{dong2014knowledge}.  Knowledge base completion aims at
solving the incompleteness problem by predicting missing links based on
existing ones. 

The great majority of research on knowledge base completion focuses on
learning effective embeddings for entities and relations through
end-to-end optimization on a pre-defined score function.  While entity
embeddings represent entity locations in the vector space, the role
played by relation embeddings is not easy to explain. Furthermore, since
relation patterns may vary significantly in a KG, it is challenging to
model relations using a single score function.  Although KG embedding
methods offer state-of-the-art performance, they are limited in several
aspects: inadequacy in relation modeling, sensitivity to embedding
dimensions, and incremental performance improvement with provision of
negative samples. To address these shortcomings, a new method, called KGBoost,
is proposed in our work, which adopts XGBoost \citep{chen2016xgboost} 
as a binary classifier for each relation. It is applied to prediction of missing links in KGs. 

KGBoost attempts to find hard negative samples so as to train a powerful
classifier for missing link prediction. Several unique characteristics
of our work are summarized below. 
\begin{itemize}
\item A modularized design is adopted by KGBoost, where each module can
be trained separately. 
\item Instead of using a single score function, each relations is assigned a binary classifier to model the unique relation pattern in KGBoost.
\item Several different negative sampling strategies for link prediction
are explored and integrated in KGBoost. 
\end{itemize}
We conduct experiments on multiple benchmark datasets, and
demonstrate that KGBoost outperforms state-of-the-art methods across most datasets. Furthermore, as compared with models trained by
end-to-end optimization, KGBoost works well under the low-dimensional setting so as to allow a smaller model size. 

\renewcommand{\figurename}{Figure}
\renewcommand{\tablename}{Table}

\section{Motivation and Related Work}\label{sec:motivation}

\subsection{Motivation}\label{subsec:motivation}

Some KG embedding models such as TransE \citep{bordes2013translating}
and RotatE \citep{sun2019rotate} model relations as simple
transformation from head entities to tail entities in the vector space.
However, relation patterns vary even in a single KG, and fixed
compositional relation patterns in TransE and RotatE have been
challenged \citep{zhang2019quaternion}. Besides, KG embedding models are
less expressive when the embedding dimension is lower
\citep{dettmers2018convolutional}.  To increase expressiveness of
embedding models, embedding in the complex space \citep{sun2019rotate,
trouillon2016complex} or a higher dimensional space
\citep{zhang2019quaternion, tang2019orthogonal} were investigated. 

\begin{figure}[!t]
\centering
\includegraphics[width=0.6\linewidth]{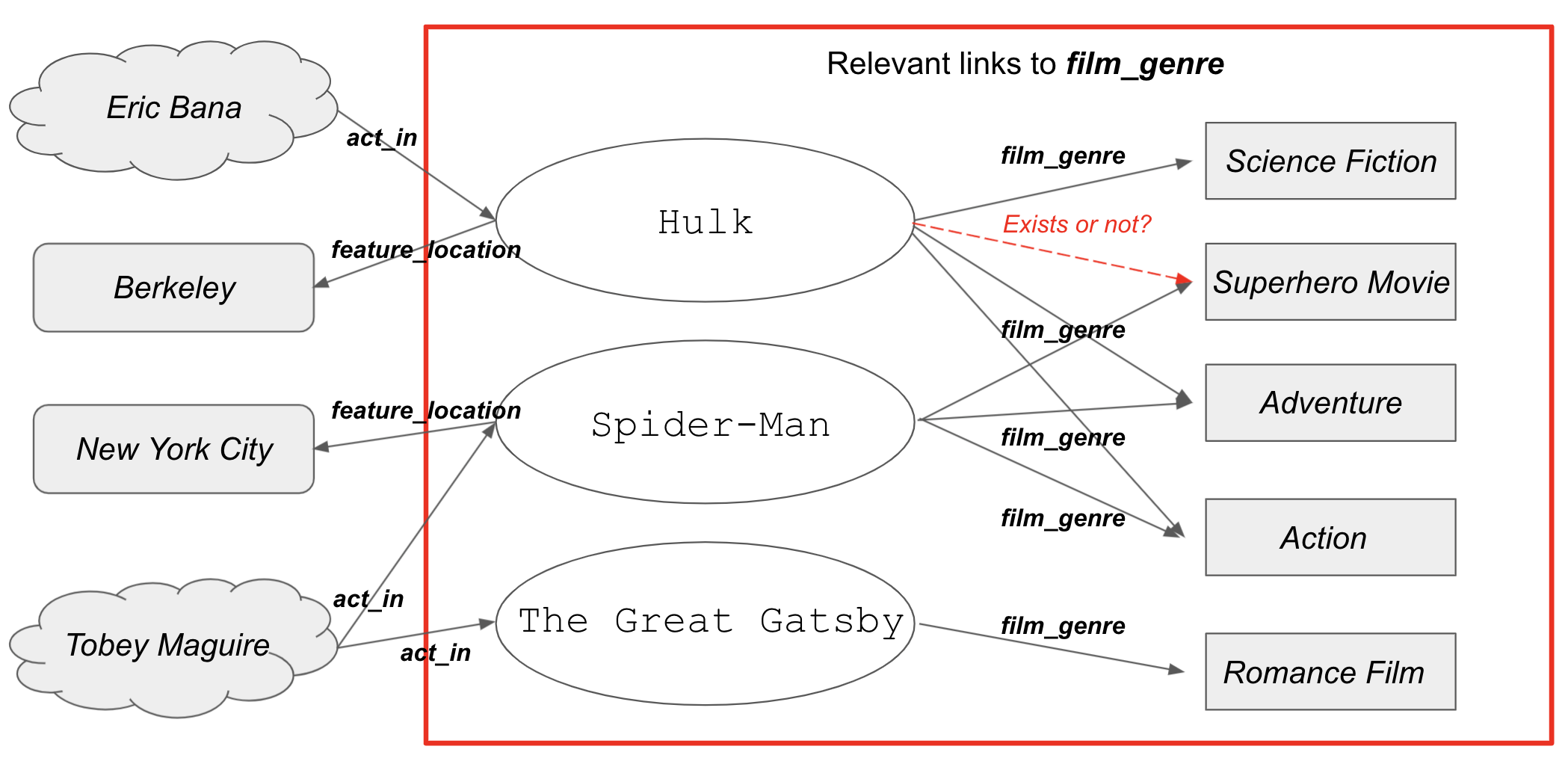} 
\caption{An example KG from Freebase.}\label{fig:kg}
\end{figure}

Given a relation, there are two possible outcomes between a head entity
and a tail entity; namely, the relation either exists or does not exist.
Thus, a relation can be potentially modeled by a binary classifier.  For
example, as in Fig. \ref{fig:kg}, ``\emph{Is Hulk a superhero movie?}''
is indeed a binary classification problem.  We may articulate whether
this is a good idea from a high-level view.  First, each relation is
modeled by an individual binary classifier and trained on links relevant
to the relation.  Second, the classifier-based approach can work well
under the low-dimensional setting since each dimension represents a
feature for the classifier. The classification performance is not
sensitive to the feature number if discriminant features already exist.
Finally, the modularized design allows KGBoost to have incremental
performance improvement with provision of negative samples. 

\subsection{Link Prediction via KG Embedding}

A great majority of existing work uses KG embedding to solve the link
prediction problem. KG embedding models are typically trained by
end-to-end optimization with a pre-defined score function $f(h, r, t)$
and a set of observed triples.  Generally speaking, KG embedding models
can be categorized into the following three types: distance-based,
semantic-matching-based and classification-based. 

Distance-based methods model entities as locations in the embedding
space and relations as linear transformation between head and tail
entities. One famous example is TransE \citep{bordes2013translating},
which models relations as translational distances and minimizes $f(h, r,
t) = || \bm{h} + \bm{r} - \bm{t} ||$, where $\bm{h}, \bm{r}, \bm{t}$ are
vectors to represent a triple $(h, r, t)$.  Although TransE can capture
compositionalities of relations, it fails to model symmetric relations
since $\bm{r} \approx \bm{0}$ for symmetric relations. TransH
\citep{lin2015learning} and TransR \citep{lin2015learning} model
symmetric relations by projecting entities to another hyperplane and
vector space respectively. While these models can capture symmetric
relations, they fail to preserve the compositional patterns. RotatE
\citep{sun2019rotate} extends embeddings to a complex space so that it
can model symmetry, asymmetry, inversion, and compositional patterns of
relations at the same time. 

Semantic matching methods \citep{bordes2014semantic} calculate the
semantic similarities among triples $(h, r, t)$. The score function is
often in form of $f(h, r, t) = \bm{h}^T \bm{M_r} \bm{t}$, where a matrix
$\bm{M_r}$ is used to model a relation. RESCAL \citep{nickel2011three}
suffers from model complexity and numerical instability caused by the
relation matrices.  DistMult \citep{yang2014embedding} confines the
relation matrices to be symmetric to reduce model complexity but it
fails to model asymmetric relations. ComplEx
\citep{trouillon2016complex} extends DistMult from the real space to the
complex space so that asymmetric relations can be modeled by the
imaginary parts of embeddings. 

Though distance- and semantic-matching-based methods are simple yet
effective, they require high-dimensional embeddings to be expressive
\citep{dettmers2018convolutional}. Therefore, classification-based
methods exploit multi-layer neural networks to increase model
expressiveness.  ConvE \citep{dettmers2018convolutional} predicts
missing links using 2D convolutional networks over entity and relation
embeddings.  InteractE \citep{vashishth2020interacte} improves the
performance by increasing feature interactions in convolutional layers.
SACN \citep{shang2019end} adopts graph convolutional network based
encoder to capture the structural information in KGs and a
convolutional-based decoder for link prediction. In this work, we adopt
relation-specific classifiers $f_r(h, t)$ to model different patterns of
relations. In addition, we adopt a modularized design to have
incremental performance improvement with provision of negative samples. 

\begin{figure*}[htbp]
\centering
\includegraphics[width=15cm]{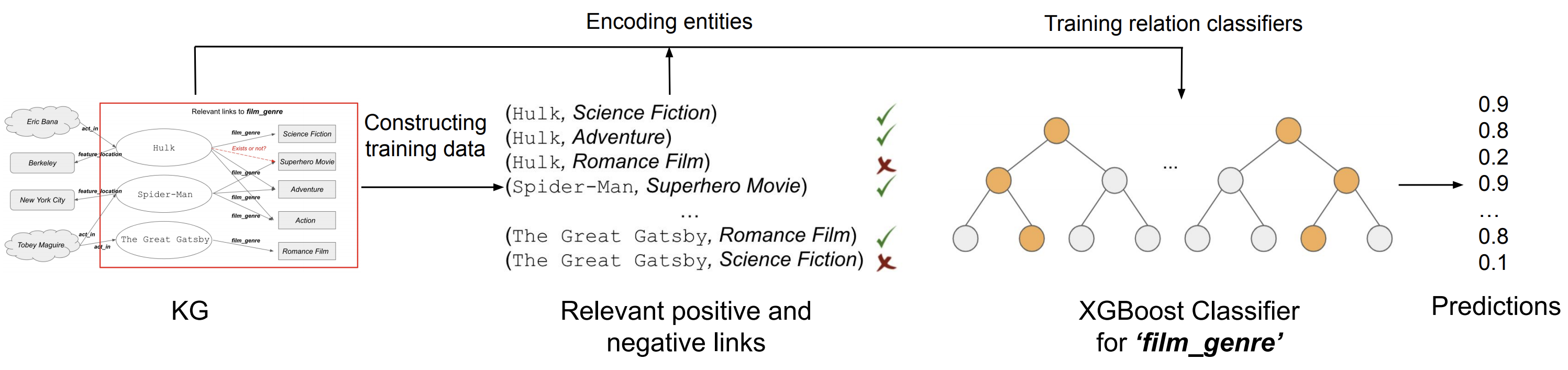}
\caption{An overview of the KGBoost pipeline.}\label{fig:pipeline}
\end{figure*} 

\subsection{Negative Sampling}

Negative sampling is important for KG applications since only observed
triples are given in KGs. The closed world assumption suggests
unobserved triples are all false. However, if KGs are incomplete under
the open word assumption, unobserved triples might be missing rather
than false.  This is more reasonable in real world applications.  Then,
an accurate choice of a negative sampling subset can contribute to the
performance of knowledge base completion models. 

Negative samples are often obtained by corrupting a true triple, $(h, r,
t)$, with a random head, $(h', r, t)$, or tail, $(h, r, t')$.  It was
proposed in \citet{wang2014knowledge} to model the probability of
corrupting heads or tails as a Bernoulli distribution to avoid false
negatives.  Adversarial learning with generative adversary networks
(GANs) for the generation of negative samples were examined in
\citet{cai2017kbgan, wang2018incorporating}. That is, effective negative
samples could be obtained through training another generator, $f_G(h',
r, t')$, simultaneously.  However, the GAN-based negative sample
generator makes the original model more complex and difficult to train.
To reduce the complexity of negative sampling, self-adversarial training
was adopted in RotatE \citep{sun2019rotate} to generate negative samples
based on the original score function $f(h',r, t')$.  In this work, we
propose a simple yet effective negative sampling strategy that is
tailored to the priors of the relations. 

\section{Proposed KGBoost Method}\label{sec:method}

We use $\mathcal{E}$ and $\mathcal{R}$ to denote sets of entities and
relations, respectively. A KG, which is represented by $\mathcal{F} =
\{(h, r, t)\} \subseteq \mathcal{E} \times \mathcal{R} \times
\mathcal{E}$, is a collection of factual triples, where $\forall h, t
\in \mathcal{E}$ and $\forall r \in \mathcal{R}$. Furthermore,
$\mathcal{G}^{(r)} = \{(h, t) \mid (h, r, t) \in \mathcal{F}\}$ is a
collection of head-tail entity pairs connected by relation $r$.  The
proposed KGBoost method consists of three main steps: (1) constructing
training data, (2) encoding entities, and (3) training relation
classifiers as shown in Fig. \ref{fig:pipeline}. They are elaborated
below.

\subsection{Constructing Training Data}\label{subsec:step1}

Constructing training data is important in KGBoost since the
quality of the training set directly affects the performance of the
classifiers. There are two criteria adopted to construct training data:
1) sufficient positive samples and 2) effective negative samples. Along
this line, we propose to incorporate inference patterns between relations to augment positive samples and generate effective negative samples based on relation priors. 

\begin{figure*}[htbp]
\centering
\includegraphics[width=15cm]{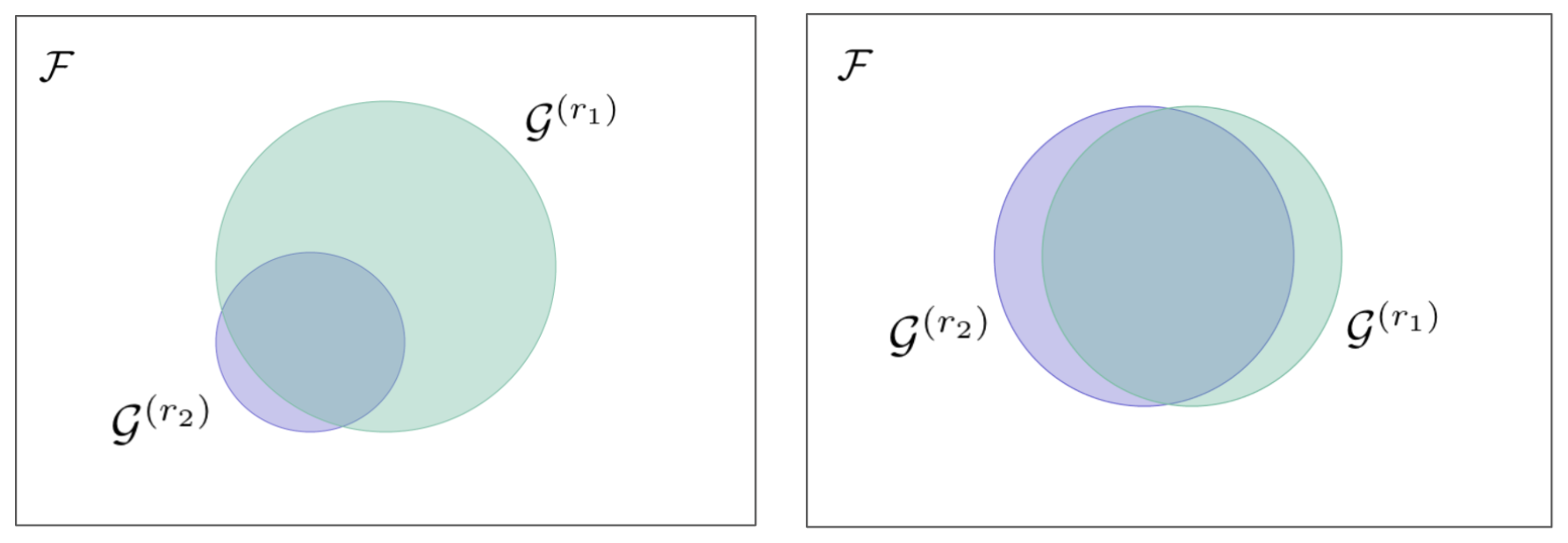}
\caption{Illustrations of subrelation scenarios.}\label{fig:subrelation}
\end{figure*}

{\bf Relation Inference} Since the classifier for each relation is trained independently, we consider inference patterns between relations to facilitate the first-order relation dependencies, i.e. subrelations. A relation,
$r_2$, is said to be a subrelation of relation $r_1$ if and only if
$\forall (h, t) \in \mathcal{G}^{(r_2)} \to (h, t) \in
\mathcal{G}^{(r_1)}$.  However, due to the incompleteness of KG, some $(h, t)$ pairs in $\mathcal{G}^{(r_2)}$ might be missing in $\mathcal{G}^{(r_1)}$. Therefore, we define an inference index to decide whether $r_2$ is a
subrelation to $r_1$ as
\begin{equation}
\textit{infer }(r_1 \mid r_2) = \frac{|\mathcal{G}^{(r_1)} \cap \mathcal{G}^{(r_2)}|} 
{|\mathcal{G}^{(r_2)}|},
\end{equation}
$r_2$ is said to be a subrelation of $r_1$ if and only if
$\textit{infer }(r_1 \mid r_2) > \delta_{sub}$, where $\delta_{sub}$ is a
threshold. We augment $\mathcal{G}^{(r_1)}$ to become $\mathcal{G}^{(r_1)}
\cup \mathcal{G}^{(r_2)}$ if $r_2$ is a subrelation of $r_1$. There are
two possible scenarios where $r_1$ can borrow positive samples from $r_2$
as shown in Fig. \ref{fig:subrelation}. 

In the first scenario, $r_2$ is a subrelation of $r_1$, but
$r_1$ is not a subrelation of $r_2$. An example is `\textit{award\_nominee}' and `\textit{award\_winner}'. If a person is the `\textit{award\_winner}' of some awards, that person is also likely to be the `\textit{award\_nominee}' for the same award. In this scenario,
$\mathcal{G}^{(r_1)}$ is augmented to be $\mathcal{G}^{(r_1)} \cup
\mathcal{G}^{(r_2)}$ but $\mathcal{G}^{(r_2)}$ stays the same. In the
second scenario, $r_2$ is a subrelation of $r_1$, and $r_1$ is
also a subrelation of $r_2$. $r_1$ and $r_2$ are either duplicate or they form a symmetric reciprocal pair, such as relation `\textit{friend\_of}' and its inverse. Then, $r_1$ and $r_2$ will
share the same positive training set $\mathcal{G}^{(r_1)} \cup
\mathcal{G}^{(r_2)}$.

{\bf Negative Sampling based on Relation Priors.} Generating negative samples
is challenging for KGs since there are only observed positive triples. na\"ive
negative sampling \citep{bordes2013translating} generates negative samples by
corrupting the tail entity (or the head entity) of an observed sample $(h,
t)$ with a random entity. na\"ive negative sampling is defined as
\begin{equation}
\mathcal{N}_{\textit{na\"ive}}^{(r)}(h, t) = \{(h, t') \mid t' \in
\mathcal{E}, (h, t') \notin \mathcal{G}^{(r)}\}.
\end{equation}

Yet, negative samples in $\mathcal{N}_{\textit{na\"ive}}^{(r)}$ does not
carry much semantics. For example, in Fig. \ref{fig:kg}, generating a
na\"ive negative sample for \texttt{Hulk} in relation
\textit{film\_genre} might yield (\texttt{Hulk}, \textit{Tobey
Maguire}), which is trivial. It does not contribute much for the models
to predict movie genres. Instead, we look for a negative sample like
(\texttt{Hulk}, \textit{Romance Film}), which is more informative than
the previous negative sample in movie genre prediction. Based on this
observation, corrupted tail entities could be drawn from different subsets of entities for different relations based on the relation priors. More specifically, only entities that have been observed in $\mathcal{G}^{(r)}$, i.e. the range of relation $r$, are considered:
\begin{equation}\label{equ:range}
\mbox{range}(r) = \{t \mid \exists h \in \mathcal{E}, (h, t)\in 
\mathcal{G}^{(r)} \}.
\end{equation}

A similar idea was mentioned in \citet{krompass2015type}, where it was
called the local-closed world assumption (LCWA). LCWA assumes that head
and tail entities of a specific relation are constrained by entity
types. The type information in KGs is often missing
\citep{huang2018knowledge} so that they use the set of existing head and
tail entities, e.g. the range, as the constraint to generate negative
samples.  An obvious drawback of negative sampling based on LCWA is that
it is likely to generate false negative samples. To mitigate this
problem, we define the co-occurrence between two tail entities as the
number of common heads in $\mathcal{G}^{(r)}$. Formally, co-occurrence between two tail entities, $t$ and $t'$ is defined as:
\begin{equation}
\mbox{co-occur}(t, t') = |\{h \mid (h, t)\in \mathcal{G}^{(r)}
\land (h, t')\in \mathcal{G}^{(r)}\}|.
\end{equation}

When the two tail entities $t$ and $t'$ are highly co-occurred, it's likely that $(h, t')$ is a false negative given $(h, t) \in \mathcal{G}^{(r)}$. Therefore, to generate a negative sample based on a positive sample, $(h, t)$, we
exclude corrupted entities $t'$ with co-occur$(t, t')$ larger than a
threshold, $\delta_{rcwc}$. As a result, the range-constrained with co-occurrence (rcwc) negative
sampling can be formulated as
\begin{equation}
\begin{split}
\mathcal{N}_{rcwc}^{(r)}(h, t) = {} & \{(h, t') \mid t' \in \textit{range(r)}, (h, t') \notin \mathcal{G}^{(r)}\} \\
                                    & \setminus \{(h, t') \mid \textit{ co-occur(t, t')}> \delta_{rcwc}\}. 
\end{split}
\end{equation}

\subsection{Encoding Entities}\label{subsec:step2}

Semantic representations for entities often carry rich information and
thus, they are suitable to be the input features for classifiers. In distance-based KG embedding models, similar entities are likely to be clustered together in the vector space due to the design of the score functions. The clustering effect provides clear decision boundaries for classifiers.
Therefore, we select two distance-based KG embedding models, TransE
\citep{bordes2013translating} in the real space and RotatE \citep{sun2019rotate} in the complex space, to encode entities. The distance functions are shown as below
\begin{itemize}
    \item \textbf{TransE} \citep{bordes2013translating} 
    \begin{equation}
       d_r(h, t) = \| \bm{h} + \bm{r} - \bm{t} \|,
    \end{equation}
    \item \textbf{RotatE} \citep{sun2019rotate}
    \begin{equation}
       d_r(h, t) = \| \bm{h} \circ \bm{r} - \bm{t} \| ^2.
    \end{equation}
\end{itemize}
The entity embeddings are then trained with the negative log likelihood loss
as in \citep{sun2019rotate}:
\begin{equation}
\begin{split}
\mathcal{L}_{emb} = {} & -\log \sigma(\gamma - d_r(h, t)) \\
                       & - \sum_{i=1}^{N} p(h_i', r, t_i') \log \sigma(d_r(h_i', t_i') - \gamma),
\end{split}
\label{equ:nll}
\end{equation}
where $\sigma(.)$ is the sigmoid function, $\gamma$ is a margin, $(h_i', r, t_i')$ is the
$i$-th negative sample, and $p(h_i', r, t_i')$ is the coefficient in self-adversarial negative sampling\citep{sun2019rotate}.

\subsection{Training Relational Classifiers}\label{subsec:step3}

Given a relation, there are two possible outcomes between a head entity
and a tail entity; namely, the relation exists or does not exist.  Thus,
a binary classifier can be used to predict the probability for a link to
exist in an entity pair. Typically, the number of negative samples is
much higher than the number of positive samples in KGs, causing the
problem of imbalanced training data. 

Ensemble and tree-based classifiers can be used to solve pairwise
matching problems and handle imbalance data \citep{furnkranz2002pairwise,
cochinwala2001efficient}. A powerful tree boosting classifier, XGBoost
\citep{chen2016xgboost}, is chosen as the relation classifier in this
work. XGBoost is a scalable gradient tree boosting system that attempts
to optimize the $k$-th tree estimator $f_k(.)$ so as to minimize
\begin{equation}\label{equ:xgboost1}
\mathcal{L}^{(k)} = \sum_{n=i}^{N} l(y_i, \hat{y}_i^{(k-1)} + f_k([\bm{h_i}; 
\bm{t_i}])) + \Omega(f_k),
\end{equation}
where $\Omega(f_k)$ is a regularization term and
$$
\hat{y_i}^{(k)} = \hat{y}_i^{(k-1)} + f_k([\bm{h_i}; \bm{t_i}])
$$
is the prediction at the $k$-th iteration.  We adopt the binary cross
entropy loss commonly used in logistic regression
\begin{equation}
\begin{split}
        l(y_i, \hat{y_i}^{(k)}) = {} & - y_i \log(\sigma(\hat{y_i}^{(k)})) \\
                                     & - (1 - y_i) \log(1 - \sigma(\hat{y_i}^{(k)}))
\end{split}
\end{equation}
The final prediction $\hat{y}_i$ becomes
\begin{equation}\label{equ:xgboost2}
\hat{y}_i = \sigma(\sum_{k=1}^{K} f_k([\bm{h_i}; \bm{t_i}])).
\end{equation}

{\bf Self-Adversarial Negative Sampling.} We investigate the strategy to provide
the classifiers easy negative samples 
in the early training stage and hard negative
samples in the later training stage with self-adversarial negative sampling,
denoted as $\mathcal{N}_{adv}$. For initial trees, we provide
a classifier with $\mathcal{N}_{\textit{na\"ive}}$ or $\mathcal{N}_{rcwc}$ to
build up its basic knowledge. As the classifier acquires some basic information,
we collect negative samples in $\mathcal{N}_{\textit{na\"ive}}$ or
$\mathcal{N}_{rcwc}$ that are mis-classified by the initial estimators to form
$\mathcal{N}_{adv}$. $\mathcal{N}_{adv}$ is used to train estimators in the later iteration to correct the mistakes made by the initial estimators. In general, $\mathcal{N}_{adv} \subseteq \mathcal{N}_{rcwc} \subseteq \mathcal{N}_{\textit{na\"ive}}$. Thus, harder negative samples are given in the training process under self-adversarial negative sampling. 

{\bf LCWA-based Prediction.} We adopt LCWA proposed in
\citet{krompass2015type} for link prediction.  When predicting missing
links for relation $r$ in the inference stage, we only consider entities
in $range(r)$. For example, we do not consider \textit{Tobey Maguire}
when predicting tails for relation \textit{film\_genre} because
\textit{Tobey Maguire} has never appeared as a movie genre. It is
worthwhile to point out that not all relations satisfy LCWA. To address
it, we further define a local-closed world (\emph{lcw}) index to check whether
a relation satisfies LCWA. To calculate $lcw(r)$, we split
$\mathcal{G}^{(r)}$ into stratified K folds, iterate through each fold,
and accumulate the number of samples that contain tail entities that do
not exist in other folds. The accumulated number is divided by the total
sample number in $\mathcal{G}_{(r)}$ to yield $1 - lcw(r)$. For triple
$(h_i, r, t_i)$, the final score function of link prediction can be
written as
\begin{equation}
    f_r(h_i, t_i) =
    \begin{cases}
        \mathbbm{1}(\{t_i \in \textit{range(r)}\})\hat{y}_i & lcw(r) > \delta_{lcw} \\
        \hat{y}_i & \text{otherwise}
    \end{cases}
\end{equation}
where $\delta_{lcw}$ is a threshold and $\mathbbm{1}\{.\}$ is an indicator function.

\section{Experiments}\label{sec:experiments}



\subsection{Experimental Setup}

{\bf Datasets.} We evaluate the proposed KGBoost method on four widely used link prediction datasets, WN18\citep{bordes2013translating}, WN18RR\citep{dettmers2018convolutional}, FB15K\citep{bordes2013translating}, FB15k-237\citep{toutanova2015observed}. The statistics of the datasets are summarized in Table.\ref{table:dataset}. WN18 and WN18RR are extracted from WordNet\citep{miller1995wordnet}, a lexical database with conceptual entities and relations. Inversed relations in WN18 are removed to form WN18RR. FB15K and FB15k-237 are extracted from Freebase\citep{bollacker2008freebase}, an instance-level knowledge base. Near-duplicate and inversed relations in FB15K are removed to form FB15k-237.

\begin{table}[!htbp]
\centering
\caption{Statistics of link prediction datasets.}
\begin{tabular}{ l l l l }
\toprule
Dataset & \#ent & \#rel & \#triples (train / valid / test) \\
\midrule
WN18      &40,943 &18 &141,442 / 5,000 / 5,000 \\
WN18RR    &40,943 &11 &86,835 / 3,034 / 3,134 \\
FB15K     &14,691 &1,345 &483,142 / 50,000 / 59,071 \\
FB15k-237 &14,541 &237 &272,115 / 17,535 / 20,466 \\
\bottomrule
\end{tabular}
\label{table:dataset}
\end{table}


{\bf Training Details.} We determine optimal hyper-parameters 
via grid search based on the MRR in the validation set with the following search values:
\begin{itemize}
\item negative sample size: 8, 16, \underline{32}, \textbf{64}.
\item number of estimators: 300, 500, \textbf{1000}, \underline{1500}.
\item max depth: \underline{3}, \textbf{5}, 7, 10.
\item learning rate: 0.01, 0.05, \textbf{\underline{0.1}}. 
\end{itemize}
Hyper-parameters to be used in datasets extracted from Freebase and WordNet are marked in bold face and with an underbar respectively. The entity embedding dimension is set to 1,000 for Fb15K and FB15k-237, and 500 for WN18 and WN18RR. We denote KGBoost with TransE as KGBoost-T and with RotatE as KGBoost-R. 
We assign an individual classifier to each relation as well as its inverse relation to have maximum flexibility.



{\bf Evaluation Metrics.} We evaluate the results using MR (Mean Rank), MRR (Mean Reciprocal Rank), and Hits@k (k=1, 3, and 10) under the filtered setting \citep{bordes2013translating}. That is, testing triples are ranked against all candidate triples that are not in training, validation, or testing set. Candidate triples are created by corrupting the tail entities in the testing triples. Since we assign classifiers for inverse relations, corrupting head entities is not required.

\subsection{Experimental Results}

Link prediction results for FB15K and WN18 are shown in Table \ref{table:main1}, and results for FB15k-237 and WN18RR are shown in Table \ref{table:main2}. KGBoost outperforms all state-of-the-art models on both datasets extracted from Freebase, FB15K and FB15k-237, since most of the relations in instance-level knowledge graphs have a fixed subset of tail entities, i.e. range. For example, the relation \textit{film\_genre} has a fixed set of tail entities that contains only 123 movie genres while there are 14,541 entities in the knowledge base. Therefore, LCWA-based link prediction can help the model to rule out most of the irrelevant candidate triples. As the irrelevant candidate triples are ruled out, \emph{rcwc} negative sampling $\mathcal{N}_{rcwc}^{(r)}$ is used to produce effective negative samples that can help the classifiers to separate the true triples from other candidate triples.

In WN18 and WN18RR, the conceptual relations don't have a fixed tail entity subset. Therefore, $\mathcal{N}_{rcwc} \approx \mathcal{N}_{\textit{na\"ive}}$ and is no longer able to generate effective negative samples. Therefore, we use $\mathcal{N}_{adv}$ to iteratively provide negative samples that have been previously mis-classified by the classifiers. KGBoost achieves the state-of-the-art performance on WN18RR and comparable results to the state-of-the-art models on WN18. 

In addition, TransE is known to have difficulties modeling symmetric relations \citep{wang2014knowledge,lin2015learning,sun2019rotate} because embeddings for symmetric relations tend to be zero vectors to minimize the score function.
This is clearly a shortcoming of modeling all triples with a single score function regardless of the relation patterns.
On the contrary, each relation is modeled by a binary classifier in KGBoost-T so the symmetric patterns could be modeled. As a result, KGBoost-T has significant improvements over TransE on FB15K, WN18, and WN18RR, which contain many symmetric and inverse relations.


\begin{table*}[t]
\centering
\caption{Link prediction results on FB15K and WN18.}
\begin{tabular}{ l  ccccc  ccccc }
\toprule
           & \multicolumn{5}{c}{FB15K} & \multicolumn{5}{c}{WN18} \\ 
           \cmidrule(l){2-6} \cmidrule(l){7-11}
           & MR & MRR & H@1 & H@3 & H@10 & MR & MRR & H@1 & H@3 & H@10  \\ 
           \midrule
TransE \citep{bordes2013translating}    & -   & 0.463 & 0.297 & 0.578 & 0.749 
           & -   & 0.495 & 0.113 & 0.888 & 0.943 \\
DistMult \citep{yang2014embedding}   & 42  & 0.798 & -     & -     & 0.893 
           & 655 & 0.797 & -     & -     & 0.946 \\
ComplEx \citep{trouillon2016complex}   & -   & 0.692 & 0.599 & 0.759 & 0.840 
           & -   & 0.941 & 0.936 & 0.945 & 0.947 \\
ConvE \citep{dettmers2018convolutional}     & 51  & 0.657 & 0.558 & 0.723 & 0.831 
           & 374 & \underline{0.943} & \underline{0.935} & \underline{0.946} & \underline{0.956} \\
RotatE \citep{sun2019rotate}    & 40  & 0.797 & \underline{0.746} & 0.830 & 0.884 
           & 309 & \textbf{0.949} & \textbf{0.944} & \textbf{0.952} & \textbf{0.959} \\
           \midrule
KGBoost-T (Ours)  & \textbf{15} & \underline{0.811} & 0.739 & \underline{0.867} & \textbf{0.915}
                  & \underline{189}& 0.820 & 0.703 & 0.936 & 0.951 \\ 
KGBoost-R (Ours)  & \underline{16} & \textbf{0.817} & \textbf{0.751} & \textbf{0.868} & \underline{0.914}
                  & \textbf{131}& 0.939 & 0.929 & \underline{0.946} & 0.955 \\
\bottomrule
\end{tabular}
\label{table:main1}
\end{table*}

\begin{table*}[t]
\centering
\caption{Link prediction results on FB15k-237 and WN18RR.}
\begin{tabular}{ l  ccccc  ccccc }
\toprule
           & \multicolumn{5}{c}{FB15k-237} & \multicolumn{5}{c}{WN18RR}     \\
           \cmidrule(l){2-6} \cmidrule(l){7-11}
           & MR & MRR   & H@1   & H@3   & H@10  & MR & MRR   & H@1   & H@3   & H@10     \\ \hline
TransE \citep{bordes2013translating}     & 357  & 0.294 & -     & -     & 0.465 
           & 3384 & 0.226 & -     & -     & 0.501    \\
DistMult \citep{yang2014embedding}   & 254  & 0.241 & 0.155 & 0.263 & 0.419 
           & 5110 & 0.43  & 0.39  & 0.44  & 0.49     \\
ComplEx \citep{trouillon2016complex}    & 339  & 0.247 & 0.158 & 0.275 & 0.428 
           & 5261 & 0.44  & 0.41  & 0.46  & 0.51     \\
ConvE \citep{dettmers2018convolutional}    & 244  & 0.325 & 0.237 & 0.356 & 0.501 
           & 4187 & 0.43  & 0.40  & 0.44  & 0.52     \\ 
RotatE \citep{sun2019rotate}    & 177  & 0.338 & 0.241 & 0.375 & 0.533 
           & 3340 & \underline{0.476} & \underline{0.428} & \underline{0.492} & \textbf{0.571}    \\
SACN \citep{shang2019end}      & -    & 0.35  & 0.26  & 0.39  & 0.54
           & -    & 0.47  & 0.43  & 0.48  & 0.54     \\ 
InteractE \citep{vashishth2020interacte}      & 172    & 0.354  & 0.263  & -  & 0.535
           & 5202    & 0.463  & 0.43  & -  & 0.528     \\ 
           \midrule
KGBoost-T (Ours)  & \underline{78}   & \textbf{0.426} & \underline{0.335} & \textbf{0.462} & \textbf{0.608} 
                  & \textbf{2405} & 0.265 & 0.062 & 0.445 & 0.544    \\
KGBoost-R (Ours)  & \textbf{77}   & \underline{0.425} & \textbf{0.336} & \underline{0.460} & \underline{0.606} 
                  & \underline{2476} & \textbf{0.478} & \textbf{0.436} & \textbf{0.493} & \underline{0.560}    \\ \hline
\end{tabular}
\label{table:main2}
\end{table*}

\begin{figure}[!htbp]
\centering
\includegraphics[width=0.85\linewidth]{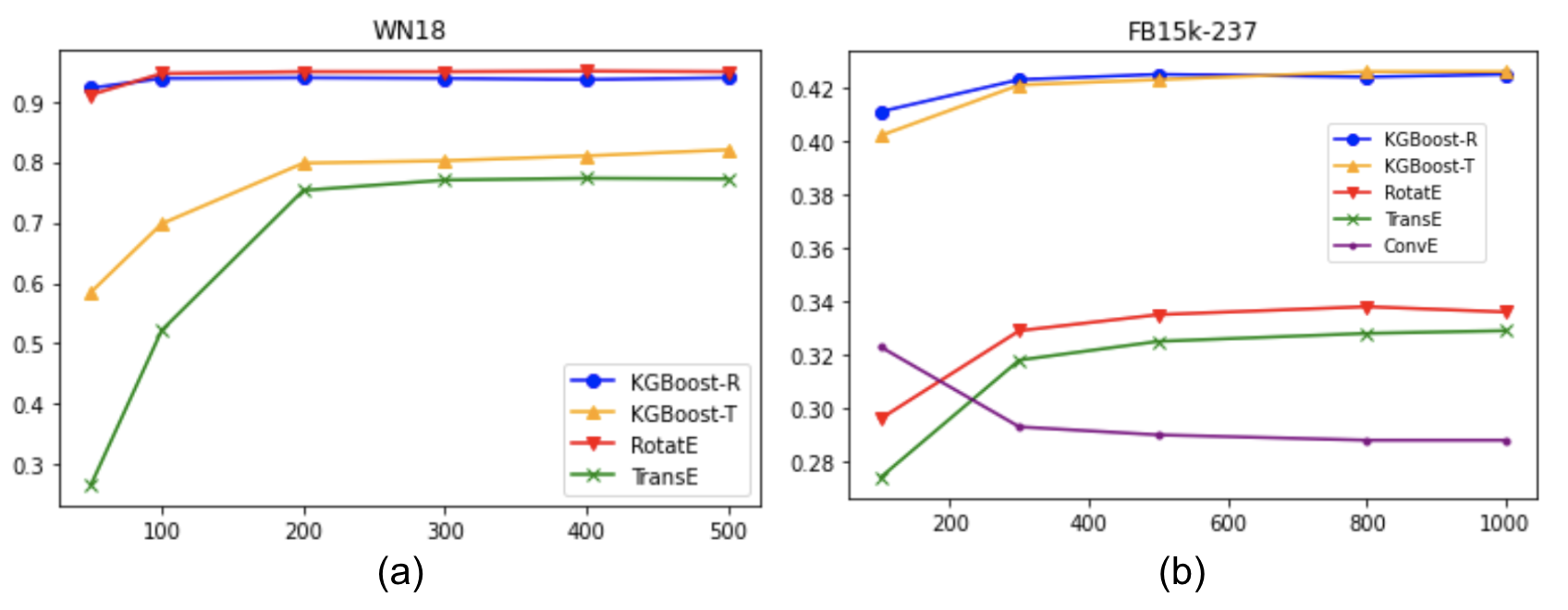}
\caption{The performance curves as a function of the embedding dimension 
for (a) WN18 and (b) FB15k-237 in MRR.}\label{fig:dim}
\end{figure}

{\bf Embedding Dimension and Performance.} As pointed out in \citet{dettmers2018convolutional}, distance-based KG embedding models, such as TransE and RotatE, require high embedding dimension for model expressiveness. However, with the modularized design, KGBoost can reach similar performance under low- and high-dimensional settings. We evaluate the performance of different models under different dimension settings in Fig. \ref{fig:dim}. 

In WN18, KGBoost-R performs slightly worse than RotatE when the embedding dimension is high. However, under the low-dimensional setting ($d = 50$) KGBoost-R is able to slightly outperform RotatE. For TransE and KGBoost-T, performance for TransE starts to drop significantly when $d < 200$, while the performance degradation for KGBoost-T is not as severe as TransE. 

In FB15k-237, both TransE and RotatE performance drop significantly under low-dimensional setting $(d = 100)$. However, KGBoost-T and KGBoost-R remain nearly the same performance as under high-dimensional settings. For ConvE, since increasing embedding dimension will result in less interactions between features in the 2D convolutional kernels, ConvE performs worse when the embedding dimension is high. The results demonstrate that KGBoost is less sensitive to embedding dimension than previous methods.

{\bf Negative Sampling and Self-Adversarial Training.} We investigate how different negative sampling strategies could affect the performance for RotatE and KGBoost-R in Table \ref{table:integration}. $\mathcal{N}_{adv}$ indicates self-adversarial setting is adopted. Self-adversarial settings have different definitions in RotatE and KGBoost. In RotatE, self-adversarial training \citep{sun2019rotate} assigns higher weights to the negative samples that have smaller margins to the positive samples in the loss function. On the other hand, in KGBoost, self-adversarial negative sampling identify previously mis-classified negative samples and correct them when training boosting trees in the later iterations. Despite the different definitions, they both aim to gradually provide harder negative samples during training. When the models are trained with na\"ive negative samples, self-adversarial settings are able to correct previous mistakes by giving more emphasis on borderline cases and boost the performance for both models.

\begin{table}[!htbp]
\centering
\caption{Model performance for FB15k-237 under different negative sampling settings.}
\begin{tabular}{lcccc}
\toprule
& \multicolumn{2}{c}{RotatE} & \multicolumn{2}{c}{KGBoost-R} \\
\cmidrule(l){2-3} \cmidrule(l){4-5}
& MRR & H@10 & MRR & H@10 \\
\midrule
$\mathcal{N}_{\textit{na\"ive}}$ & 0.295 & 0.480 & 0.307 & 0.479 \\
$\mathcal{N}_{\textit{na\"ive}}$ + $\mathcal{N}_{\textit{adv}}$ 
                                 & \textbf{0.338} & \textbf{0.533}     & 0.354 & 0.532 \\

$\mathcal{N}_{\textit{rcwc}}$    & 0.248 & 0.419 & \textbf{0.425}& \textbf{0.606} \\
$\mathcal{N}_{\textit{rcwc}}$ + $\mathcal{N}_{\textit{adv}}$ 
                                 & 0.218 & 0.380     & 0.424 & \textbf{0.606}  \\ 

           
\bottomrule
\end{tabular}
\label{table:integration}
\end{table}

When the models are trained with \emph{rcwc} negative samples, which are consider to carry more semantics than na\"ive negative samples, lack of trivial negative samples causes the poor performance for RotatE. On the other hand, since KGBoost uses pre-trained entity embeddings as input, trivial negative samples are widely separated from positive samples in the embedding space and no longer required. The modularized design of KGBoost allows to have incremental performance improvement with provision of effective negative samples.

{\bf Ablation Study.} We evaluate how each module in KGBoost affects the performance in Table \ref{table:ablation}. Relation inference is incorporated to facilitate first-order dependencies between relations and is able to boost the performance from 0.327 to 0.425 for FB15k-237 and 0.469 to 0.478 for WN18RR in MRR. \emph{rcwc} negative sampling incorporates relation priors to generate negative samples with semantics. It is able to boost the performance from 0.307
to 0.425 for FB15k-237 and 0.475 to 0.478 for WN18RR in MRR. LCWA-based prediction filters out irrelevant candidate triples during testing. It boosts the performance from 0.219 to 0.425 for FB15k-237 and 0.476 to 0.478 for WN18RR in MRR.

In general, KGBoost performs better on instance-level knowledge graphs, such as Freebase, than knowledge base with conceptual entities and relations, such as WordNet, because different relations in instance-level knowledge graphs have different priors, e.g. relation ranges. KGBoost is able to make specific prediction for each relation tailored to relation priors. 

\begin{table}[!htbp]
\centering
\caption{Ablation study evaluated in MRR.}
\begin{tabular}{l c c }
\toprule
& WN18RR & FB15k-237  \\ \cmidrule{2-3}
\makecell[l]{Complete KGBoost} & 0.478 & 0.425 \\
\midrule
\makecell[l]{w.o. relation inference} & 0.469 &\makecell{0.327} \\ 
\midrule
\makecell[l]{w.o. \emph{rcwc} negative sampling} & 0.475 &\makecell{0.307} \\ 
\midrule
\makecell[l]{w.o. LCWA-based prediction} & 0.476 & \makecell{0.219} \\
\bottomrule
\end{tabular}
\label{table:ablation}
\end{table}

\section{Conclusion and Future Work}\label{sec:conclusion}

In this paper, we propose KGBoost, a knowledge base completion method with a modularized design to model unique pattern of each relation. Different from previous KG embedding models using a single score function for all relations, we formulate link prediction in each relation as a binary classification problem and leverage XGBoost to predict missing links. Besides, range-constrained with co-occurrence (rcwc) negative sampling and self-adversarial negative sampling are proposed to generate effective negative samples. Experimental results show that KGBoost not only outperforms state-of-the-art methods in link prediction, but also works well under low-dimensional setting.

In the future, we aim to extend KGBoost to predict missing links for emerging entities and relations. Since KGs are constantly evolving, new entities and relations are introduced to the knowledge base frequently. When a new entity or relation is added, existing KG embedding models need to be re-trained on the entire KG again. In KGBoost, each relation classifier is trained separately and entity embeddings can be pre-trained. As a result, KGBoost has the potential to handle emerging entities and relations and can be extended to an inductive setting.

\bibliographystyle{plainnat}
\renewcommand\refname{Reference}
\bibliography{references}

\end{document}